\documentclass{article}
\usepackage{iclr2018_conference,times}
\usepackage{hyperref}
\usepackage{url}
\usepackage[T1]{fontenc}
\usepackage{amssymb,amsmath,amsthm, amsfonts}
\usepackage{amssymb}
\usepackage{float}
\usepackage{graphicx}
\usepackage{algorithm}
\usepackage{xcolor,soul}

\title{Distributional Multivariate Policy Evaluation and Exploration with the Bellman GAN}

\author{
Dror Freirich$^{*}$ and Ron Meir\\
The Viterbi Faculty of Electrical Engineering\\
Technion - Israel Institute of Technology\\
\and 
{\bf Aviv Tamar}\\Berkeley AI Research Lab\\UC Berkeley
}

\iclrfinalcopy % Uncomment for camera-ready version, but NOT for submission.

\DeclareMathOperator*{\argmax}{argmax}

\begin{document}

\maketitle

\begin{abstract}
The recently proposed distributional approach to reinforcement learning (DiRL) is centered on learning the distribution of the reward-to-go, often referred to as the value distribution. In this work, we show that the distributional Bellman equation, which drives DiRL methods, is equivalent to a generative adversarial network (GAN) model. In this formulation, DiRL can be seen as learning a deep generative model of the value distribution, driven by the discrepancy between the distribution of the current value, and the distribution of the sum of current reward and next value. We use this insight to propose a GAN-based approach to DiRL, which leverages the strengths of GANs in learning distributions of high-dimensional data. In particular, we show that our GAN approach can be used for DiRL with multivariate rewards, an important setting which cannot be tackled with prior methods. The multivariate setting also allows us to unify learning the distribution of values and state transitions, and we exploit this idea to devise a novel exploration method that is driven by the discrepancy in estimating both values and states.
\end{abstract}

\section{introduction}

Deep reinforcement learning (DRL) has been applied to a wide range of problems in robotics and control, where policies can be learned
according to sensory inputs without assuming a model of the environment
\citep{mnih2015human,schulman2015trust}. Until recent years,
most RL methods have relied on estimating the expected future return, a.k.a.~the `value function', 
for carrying out the next action. 
Recent studies \citep{bellemare2017distributional,dabney2017distributional} have
suggested that a Distributional Reinforcement Learning (DiRL) approach, where the value distribution, rather than the expectation are learned, leads to improved learning performance\footnote{While the value is customarily defined as an expectation over a random variable, the return in \eqref{eq:return-def}, we follow \cite{bellemare2017distributional} and use the term value distribution for the distribution of the return.}. 

DiRL algorithms such as C$51$~\citep{bellemare2017distributional} and Quantile Regression DQN \citep{dabney2017distributional} learn a mapping from states to a parametric distribution over the return, based on a distributional Bellman equation. Empirically, the distributional perspective to RL has been shown to significantly improve performance on challenging benchmarks \citep{bellemare2017distributional,hessel2017rainbow}.

In this work, we provide a new approach to DiRL, building on an equivalence between the distributional Bellman equation and Generative Adversarial Networks (GANs) \citep{goodfellow2014generative,arjovsky2017wasserstein}. From this perspective, DiRL can be seen as learning a deep generative model of the value distribution, driven by the discrepancy between the distribution of the current value, and the distribution of the sum of current reward and next value. This view allows us to leverage GAN techniques for improving DiRL algorithms. In particular, GANs are known to be effective models for high-dimensional and correlated data such as images. We exploit this fact to develop a DiRL method for multivariate rewards, a setting for which previous DiRL methods are not suitable. 

Multivariate rewards are important for domains where the natural performance evaluation of a policy depends on several different factors that cannot be easily combined into a single reward scalar~\citep{vamplew2011empirical}. Here, we also show that the multivariate reward approach allows us to unify learning the distribution of future rewards and states under a common learning framework. We use this combined framework to develop a novel exploration strategy for RL, where the signal driving the exploration is the discrepancy in the learned distribution of states and rewards. We demonstrate the efficiency of our methods on high-dimensional test-benches.

In summary, our specific contributions in this work are:
\begin{enumerate}
\item Demonstrate an equivalence between the distributional Bellman equation and GANs, where we consider the Wasserstein metric \citep{arjovsky2017wasserstein,villani2008optimal} as a distributional distance measure.
\item Introduce the multivariate distributional Bellman equation and demonstrate policy evaluation
on vector-valued reward functions.
\item Establish a novel reward-systematic exploration scheme,
combining exploration with transition model learning.
\end{enumerate}

\section{Preliminaries and formulation}

{Let $(\mathcal{S},\mathcal{A},\mathcal{{R}},\mathcal{P},\gamma)$ be a Markov Decision Process (MDP; \citealt{bertsekas2005dynamic}), where $\mathcal{S}$ is the state-space, $\mathcal{A}$ is the action space and $\mathcal{R}$ is the space of possible rewards. The states transition according to the \emph{unknown} rule, $s_{t+1}\sim\mathcal{{P}}(\cdot|s_{t},a_{t})$ and $\gamma\in[0,1)$ is the discount factor. The action $a_{t}$ is drawn w.r.t.~some policy
$\pi(\cdot|s_{t})$.} We discuss the case of full state observations
for simplicity. A reward function $r_{t}= r(s_{t},a_{t},s_{t+1})\in\mathcal{{R}}$
is supplied by the environment and may be unknown. By $\mathcal{Z}$ we
denote the set of transformations from state-action space to probability
distributions over rewards, $\mathcal{Z}:\mathcal{S}\times\mathcal{A}\rightarrow\mathcal{{P}}(\mathcal{{R}})$.
Given a state $s$ and an action $a$, we consider
the (random) return, when starting from $s$ and taking action $a$
\begin{equation}\label{eq:return-def}
    R(s,a)\triangleq \sum_{t=0}^\infty \gamma^{t}r_{t}, \mathrm{\quad  s.t. \; } s_{0}=s,a_{0}=a.
\end{equation}
Following \cite{bellemare2017distributional}, we use the notation $Z^{\pi}(s,a)$ for the return, and $Z^{\pi}(s,a)\in\mathcal{{Z}}$ to imply that $Z^{\pi}(s,a)$ is distributed according to a distribution belonging to the set of distributions $\mathcal{{Z}}$. Since we will be assuming a fixed policy for most of this work, we will simply use $Z$, rather than $Z^\pi$, to denote the random return. 

\subsection*{Notation}

Throughout this paper we use the following acronyms. MDP - Markov Decision Process, 
RL - Reinforcement Learning, NN - Neural Network, GAN - Generative Adversarial Network,
DQN - Deep Q-Network. $W_{p}$ is the Wasserstein-$p$ metric . For any two random variables
$X$ and $Y$, $X=Y$ denotes equality in distribution. By $\|\cdot\|$ we denote
the $l_{2}$-norm, and by $I_{m}$ the identity matrix of dimension $m$. $\delta_{A}$ is the indicator function of the set $A$.

\subsection{Distributional RL}\label{subsec:Distributional-perspective-on}
Given an initial state $s$ and an action $a$, the $Q$-function is defined as the expectation

\begin{equation}
Q^{\pi}(s,a)\triangleq \mathbb{{E}}_{\pi}R(s,a)=\mathbb{{E}}_{a_{t}\sim\pi(s_{t})}  \left[ \sum_{t\geq0}\gamma^{t}r_{t} | {s_{0}=s,a_{0}=a} \right].
\end{equation}
A useful property of $Q$-functions is that they obey the Bellman equation \citep{bertsekas2005dynamic}, 
\begin{equation}
Q^{\pi}(s,a)=\mathbb{{E}}_{\pi}[r(s,a)+\gamma Q^{\pi}(s',a')],
\end{equation}
where $s'\sim \mathcal{{P}}(\cdot|s,a)$, and $a'\sim \pi(\cdot|s')$.

An \textit{optimal policy} is a policy satisfying $\pi^{*} \in \argmax_{\pi}\mathbb{{E}}_{\pi}Q(s,\pi(\cdot|s))$
for all $s\in\mathcal{{S}}$. The $Q$-function of an optimal
policy, denoted $Q^{*}$, satisfies the optimality equation
\begin{equation}\label{eq::preliminaries::bellman_optimality_eqn}
Q^{*}(s,a)=\mathbb{{E}}[r(s,a)+\gamma \max_{a'}Q^{*}(s',a')].
\end{equation}
In RL we typically try to maximize the expected cumulative reward, namely the $Q$-function. Therefore,
many RL approaches involve approximating solutions of \eqref{eq::preliminaries::bellman_optimality_eqn}
\citep{bertsekas1995dynamic,mnih2015human}.

While the goal in RL is to maximize the expected return, it has recently been observed that learning the full distribution of the return $Z(s,a)$, rather than the expectation $\mathbb{E}\left[Z(s,a)\right]$, leads in practice to better performing algorithms~\citep{bellemare2017distributional,hessel2017rainbow}. 

For learning the distribution of $Z(s,a)$ under a fixed policy $\pi$, \citet{bellemare2017distributional} showed that the\textit{ Bellman operator}, 
\begin{equation}
\mathcal{T^{\pi}}Z(s,a)\triangleq r(s,a)+\gamma Z(s',a'),
\end{equation}
is a $\gamma-$contraction over the distribution space $\mathcal{{Z}}$
under the metric 
\begin{equation}
\bar{d}_{p}=\sup_{s,a} W_{p}(Z_{1}(s,a),Z_{2}(s,a)),\qquad Z_{1},Z_{2}\in\mathcal{{Z}}.
\end{equation}
Here again, $s'$ is a random variable drawn from $\mathcal{{P}}(\cdot|s,a)$ and 
$a'\sim \pi(\cdot|s')$, and $W_{p}$ is the Wasserstein-$p$ metric. 

For practical reasons, DiRL algorithms use a parametric family of
distributions $Z_{\theta}(s,a)$ to approximate $Z(s,a)$. Previous work explored using deep neural networks to map from $s,a$ to either a distribution defined by a fixed set of particles~\citep{bellemare2017distributional}, or a mixture of uniform distributions~\citep{dabney2017distributional}. An approximation of the Bellman operator was used to update the distribution parameters.

\subsection{Generative Adversarial Networks} \label{sec:GANs}

GANs train two competing models (typically NNs) \citep{goodfellow2014generative}. The \textit{generator} takes noise
$z\sim P_{z}$ as input and generates samples according to some transformation,
$G_{\theta}(z)$. The \textit{discriminator} takes samples from
both the generator output and the training set as input, and aims to distinguish between the input sources. During the training process, the generator learns to produce samples similar to real data. The discriminator, on the other hand, learns to better distinguish generated samples from real ones. The networks are trained until ideally reaching equilibrium,
where the generated samples are indistinguishable from real ones.

In their original work, \cite{goodfellow2014generative} measured discrepancy
between the generated and the real distribution using the Kullback-Leibler
divergence. However, \cite{arjovsky2017wasserstein} demonstrated sequences of simple probability distributions that do not converge under this distance. It was suggested that considering the Wassertein-$1$ distance leads to enhanced generated sample quality and improved stability under simpler NN architectures. 
Wasserstein-GANs exploit the Kantorovich-Rubinstein duality \citep{villani2008optimal}
\begin{equation}
W_{1}(\mathbb{P}_r,\mathbb{P}_g)=\sup_{f \in 1-\mathrm{Lip}}\left\{ \mathbb{E}_{x \sim \mathbb{P}_r}f(x)-\mathbb{E}_{x \sim \mathbb{P}_g}f(x)\right\}, \label{eq:Pre::Wasserstein::duality}
\end{equation}
where $1-\mathrm{Lip}$ is the class of Lipschitz functions with Lipschitz constant $1$, in order to approximate the distance between the real distribution , $\mathbb{P}_r$, and the generated one, $\mathbb{P}_g$. The GAN objective is then to train a generator model $G_\theta (z)$ with noise distribution $P_z$ at its input, and a critic $f\in 1-\mathrm{Lip}$, achieving
\begin{equation}\label{eq:Pre:WGAN}
\min_{G_{\theta}(\cdot)}\max_{f \in 1-\mathrm{Lip}}\left\{ \mathbb{E}_{z \sim \mathbb{P}_r}f(x)-\mathbb{E}_{z \sim P_z}f(G_{\theta}(z))\right\}.
\end{equation}
The WGAN-GP \citep{gulrajani2017improved} is a stable training algorithm for \eqref{eq:Pre:WGAN} that employs stochastic gradient descent and a penalty on the norm of the gradient of the critic with respect to its input. The gradient norm is sampled at random points along lines between a real sample, $x$, and a generated sample, $G_\theta (z)$, 
\begin{equation}
\tilde{x} = \varepsilon x +(1-\varepsilon)G_\theta (z), \quad x\sim\mathbb{P}_r,\ z\sim P_z,\ \varepsilon \sim U[0,1].
\end{equation}
The penalty term $\lambda \mathbb{E}_{\tilde{x}} \left(\|\nabla_{\tilde{x}} f(\tilde{x})\|-1\right)^2$ is then added to the discriminaor optimization objective, where $\lambda$ is the gradient penalty coefficient. By employing  this penalty, the norm of the critic gradient is penalized for deviating from $1$, thus encouraging $f$ to belong to $1-\mathrm{Lip}$.

\subsection{Intrinsic-reward based exploration}\label{ssec:exploration_background}

Exploration is a key challenge in RL. While efficient algorithms and performance guarantees are available for the model-based setting with finite state and action spaces (e.g., \cite{kearns2002near,osband2013more,tewari2008optimistic}), the situation is very different in the model free setting, and, in particular, for large state and action spaces. Within a model-free setting, the unknown environment and rewards are not directly modeled, and actions are learned by the agent through trial-and-error experience. At each stage of the process an agent must balance exploration and exploitation. Exploitation uses the knowledge gained so-far about the environment, and attempts to use this to maximize the return. Since the knowledge gained is always partial and approximate, exploration is required in order to improve the agent's knowledge about environment, thereby improving future exploitation. Exploration is particularly difficult in a model free setting, where the environment itself is not modeled. Simple exploration techniques in model-free RL 
draw randomized actions (e.g. $\epsilon$-greedy, Boltzmann exploration \citep{sutton1998reinforcement}), or mildly perturb policy improvements \citep{lillicrap2015continuous}. 

A promising recent approach to exploration in model free settings uses the notion of curiosity and internal reward \citep{oudeyer2007intrinsic,schmidhuber2010formal} in order to direct the learner to regions in state-action space where system uncertainty is large. Such methods aim to explore regions in state-action space through actions that lead to poorly predictable outcomes, namely to `surprise'. 

These methods often set a trade-off between exploitation and exploration using a tuning parameter $\eta$ and a combined reward function, 
\begin{equation}\label{eq:intrinsic}
r'(s_{t},a_{t},s_{t+1})=r(s_{t},a_{t})+\eta r^{i}(s_{t},a_{t},s_{t+1}).
\end{equation}
In this formulation, intrinsic rewards $r^{i}(s_{t},a_{t},s_{t+1})$ measure information gains about the agent's internal belief of a dynamical model it holds. For example, \citet{houthooft2016vime} capture information gain through the notion of mutual information which is approximated and estimated by their algorithm. 

\section{Related Work}

The C$51$ algorithm of \cite{bellemare2017distributional} represented the value distribution using a fixed set of $51$ particles, and learned the probability of each particle. \citet{dabney2017distributional} extended this approach to particles with adjustable locations, where each particle corresponds to a fixed uniform distribution. Both approaches, which rely on `discretization' of the value distribution, do not scale to high-dimensional multivariate reward distributions.

Concurrently, and independently from us, \citet{doan2018gan} showed a similar equivalence between the distributional Bellman equation and GANs, and used it to develop a GAN Q-learning algorithm. Compared to that work, which did not show any significant improvement of GAN Q-learning over conventional DiRL methods, we show that the GAN approach can be used to tackle multivariate rewards, and use it to develop a novel exploration strategy. 

In the context of model based RL, \citet{asadi2018equivalence} showed an equivalence between Wasserstein and value-aware methods for learning state transition models. This is different from our framework, that is able to learn both the state transitions and the value distribution using a single deep generative model.

\citet{tang2018exploration} proposed a distributional
RL based exploration strategy with Bayesian parameter updates. This approach introduces a minimization objective combining the expected distributional discrepancy from observed data together with an exploration term, encouraging a high-entropy distribution. In our work, we suggest that the learning of a distributional discrepancy by itself naturally yields a motivation for exploration.

In the context of intrinsic-reward based exploration, the VIME method of \citet{houthooft2016vime} described above, does not take information about the distribution of incoming rewards into consideration. This may become an advantage
where rewards are sparse, but misses crucial information about the task elsewhere. 
In another approach, \cite{pathak2017curiosity} also build on the curiosity driven learning paradigm, and propose an algorithm that is based on jointly learning forward and inverse models over feature representations, and using the prediction error to augment the external reward (if the latter exists) as in \eqref{eq:intrinsic}. An interesting aspect of both these, and other curiosity based approaches, is that they can learn without any external reward.

Other recent work in RL exploration that is not directly related to curiosity, and does not aim to learn the environment, is based upper confidence bounds, developed originally for bandits, and later extended to model based RL. Recently, \cite{pazis2016efficient} have used this framework in the context of model free Q-learning, where the standard Q-function is penalized in such a way that unvisited states are visited more often.

\section{Equivalence between DiRL and GANs}\label{sec:methods}

In this section, we show that the distributional Bellman equation can be interpreted as a GAN, leading to a novel approach for learning the return distribution in DiRL. Given a fixed stochastic policy $\pi$, where $a_{t}\sim\pi(\cdot|s_{t})$, we construct a state-action conditioned generative model with an output distribution $Z(s,a)\in\mathcal{{Z}}$ such that,
\begin{equation}
Z(s,a)\overset{\bar{d}_{1}}{=}T^{\pi}Z(s,a),\label{eq:bellmanDistEqn}
\end{equation}
where
\begin{equation}
T^{\pi}Z(s,a)\triangleq r(s,a,s')+\gamma Z(s',a'),\quad s'\sim \mathcal{P}(\cdot|s,a), \quad a'\sim\pi(\cdot|s')~.
\end{equation}
The notation \eqref{eq:bellmanDistEqn} implies that $\overline{d_{1}}\triangleq \sup_{s,a}W_{1}(Z(s,a),T^{\pi}Z(s,a))$ vanishes.

We can
solve \eqref{eq:bellmanDistEqn} via a game formulation: due to the
Kantorovich-Rubinstein duality \eqref{eq:Pre::Wasserstein::duality} we have a 1-Lipschitz function (in
its first argument), $f(r|s,a)$, s.t. 
\begin{equation}\label{eq:Kantor}
W_{1}(Z(s,a),T^{\pi}Z(s,a))=\mathbb{{E}}_{r\sim T^{\pi}Z(s,a)}f(r|s,a)-\mathbb{{E}}_{r\sim Z(s,a)}f(r|s,a),\,\forall s,a.
\end{equation}

\begin{figure}[H]
\centering{}
\begin{tabular}{cc}
\includegraphics[width=0.5\textwidth]{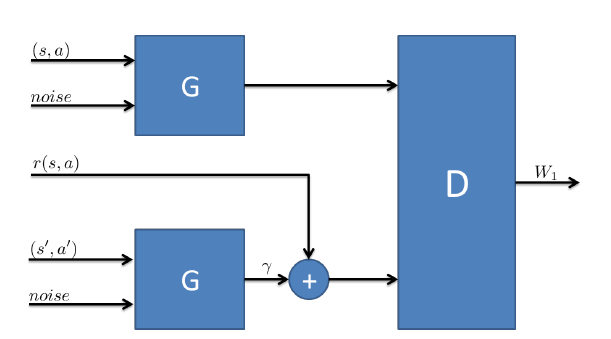} & \includegraphics[width=0.5\textwidth]{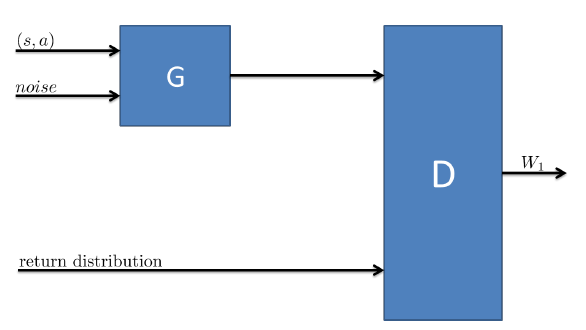}\tabularnewline
(a)  & (b)\tabularnewline
\end{tabular}\emph{\caption{GAN configurations.
(a) Bellman-GAN; (b) WGAN~\citep{arjovsky2017wasserstein}}\label{fig:bellGAN:Configurations}}
\end{figure}

Similarly to \cite{arjovsky2017wasserstein}, we make use of this duality
for training a \emph{conditional GAN} in a novel configuration we call
a \textit{Bellman-GAN} (Figure \ref{fig:bellGAN:Configurations}(a)).
The generator $G_{\theta}(\cdot|s,a)$ is a generative NN model with
parameters $\theta$, whose input consists of random noise with distribution
$P_{z}$, and whose output distribution imitates that of $Z^{\pi}(s,a)$. The discriminator
learns a critic $f_{\omega}(\cdot|s,a)$ (we will sometimes omit the
state-action condition for simplicity) . The optimization objective is

\begin{subequations} \label{eq:bellGAN}
\begin{align}
\min_{G_\theta(\cdot|s,a)}\max_{f_\omega(\cdot|s,a)\in1-\mathrm{Lip}}&\mathcal{{L}}_{\pi}(G,f), \label{eq:bellGAN::gameFormula}\\
&\mathcal{{L}}_{\pi}(G,f)=\mathbb{{E}}_{z\sim P_{z},a_{t+1}\sim\pi(\cdot|s_{t+1})}\Lambda(G_{\theta},f_{\omega}))|_{(z,s_{t},a_{t},r_{t},s_{t+1},a_{t+1})}.\label{eq:bellGAN::lossDefinition}
\end{align}
\end{subequations}
Here, $z\sim P_{z}$ is the input noise and $\Lambda$ is defined by 
\begin{equation}\label{eq:Lambda-def}
\Lambda(G_{\theta},f_{\omega})|_{(z,s,a,r,s',a')} \triangleq f_{\omega}(r+\gamma G_{\theta}(z|s',a'))-f_{\omega}(G_{\theta}(z|s,a)).
\end{equation}

\emph{}

We emphasize that unlike the conventional WGAN~\citep{arjovsky2017wasserstein}, illustrated in  Figure \ref{fig:bellGAN:Configurations}(b), in the Bellman GAN (Figure \ref{fig:bellGAN:Configurations}(a)), both distributions are generated and there is no `real data' distribution. 

We train our critic using the WGAN-GP scheme \citep{gulrajani2017improved},
with a penalty factor of $\lambda$ (see section \ref{sec:GANs}). We term our algorithm Value Distribution Adversarial Learning (VDAL); pseudo-code is provided in Algorithm \ref{alg:Distributional-Value-Learn}.

\begin{algorithm}[h]
\caption{\label{alg:Distributional-Value-Learn} Value Distribution Adversarial Learning (VDAL)  }

INPUT: discriminator parameters $\omega$, generator parameters $\theta$, fixed policy $\pi(\cdot|s)$, discount factor $\gamma$.

PARAMETERS: number of steps $T$, learning rate $\alpha$, penalty factor $\lambda$, minibatch size $m$, $n_{critic}$ = 5. 
\begin{enumerate}
\item For T steps, act according to $a_{t}\sim\pi(\cdot|s_{t})$.
\begin{enumerate}
\item Observe $(s_{t},a_{t},s_{t+1})$ and draw reward $r_{t}$.
\item Draw the next action according to $a_{t+1} \sim \pi(\cdot|s_{t+1})$.
\item Store $(s_{t},a_{t},r_{t},s_{t+1},a_{t+1})$ in replay pool.
\end{enumerate}
\item Train critic (repeat $n_{critic}$ times)
\begin{enumerate}
\item Sample $\{ (s^{(i)}_{t},a^{(i)}_{t},r^{(i)}_{t},s^{(i)}_{t+1},a^{(i)}_{t+1}) \}_{i=1}^{m} $ from replay pool.
\item Sample both $\{ z^{(i)}\}$ and $\{z'^{(i)}\}$, $i=1,\ldots,m$, from $P_z$, and $\{ \varepsilon^{(i)}\}$, $i=1,\ldots,m$, from $U[0,1]$.
\item $x^{(i)}_\theta = G_\theta({z}^{(i)}|s^{(i)}_{t},a^{(i)}_{t}), \quad {x'}^{(i)}_\theta = r^{(i)}_{t}+\gamma 
G_\theta({z'}^{(i)}|s^{(i)}_{t+1},a^{(i)}_{t+1})$.
\item $\tilde x^{(i)} \leftarrow \varepsilon^{(i)} x^{(i)}_\theta + (1-\varepsilon^{(i)}) {x'}^{(i)}_\theta$. 
\item $g_\omega \leftarrow \frac{1}{m} \nabla_\omega \sum_{i=1}^{m} \left[ f_\omega \left( {x}^{(i)}_\theta \right) - f_\omega \left({x'}^{(i)}_\theta\right) + \lambda  \left( \| \nabla_{\tilde x} f_\omega (\tilde x^{(i)})\|-1\right)^2\right]$.
\item $\omega \leftarrow \mathrm{Adam}(\omega,g_\omega,\alpha)$.
\end{enumerate}
\item Train generator
\begin{enumerate}
\item Sample $\{ (s^{(i)}_{t},a^{(i)}_{t},r^{(i)}_{t},s^{(i)}_{t+1},a^{(i)}_{t+1}) \}_{i=1}^{m} $ from replay pool.
\item Sample both $\{ z^{(i)}\}$ and $\{z'^{(i)}\}$, $i=1,\ldots,m$, from $P_z$.
\item $x^{(i)}_\theta = G_\theta({z}^{(i)}|s^{(i)}_{t},a^{(i)}_{t}), \quad {x'}^{(i)}_\theta = r^{(i)}_{t}+\gamma 
G_\theta({z'}^{(i)}|s^{(i)}_{t+1},a^{(i)}_{t+1})$.
\item $g_\theta \leftarrow -\frac{1}{m} \nabla_\theta \sum_{i=1}^{m} \left[ f_\omega \left( {x}^{(i)}_\theta \right) - f_\omega \left({x'}^{(i)}_\theta\right)\right]$.
\item $\theta \leftarrow \mathrm{Adam}(\theta,g_\theta,\alpha)$.
\end{enumerate}
\end{enumerate}
\end{algorithm}

\subsection{Q-function and Z-distribution estimation} \label{sec::ZQsimEstim}

\begin{figure}
\centering
\includegraphics[scale=0.60]{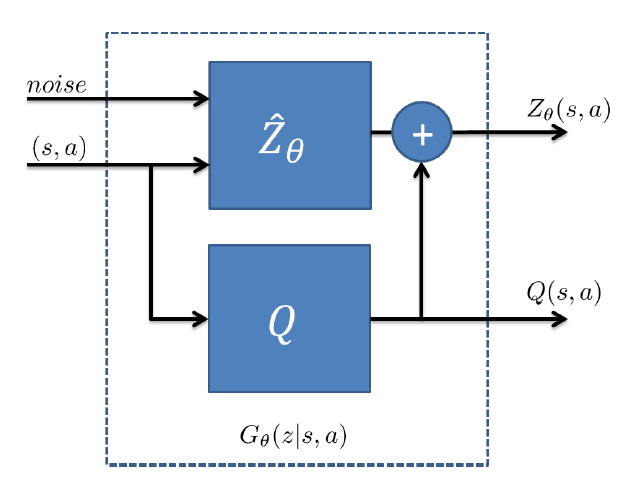}\caption{\label{fig:Simultaneous-QandZ}Simultaneous estimation of Q-function and Z-distribution}
\end{figure}

Learning the discounted reward distribution of $Z(s,a)$ at every point of state-action space may lead to slow convergence for large spaces (recall that we are concerned here with a fixed policy). However, learning the $Q$-function is a simpler task, and may be utilized at an earlier stage of training. By setting the generator architecture given in Figure \ref{fig:Simultaneous-QandZ}, we manage to estimate the $Q$ and $Z$ functions concurrently, without significant additional computational cost. Here $Q(s,a)$ is trained using DQN to satisfy
the optimality equation \eqref{eq::preliminaries::bellman_optimality_eqn}, then $G_\theta(\cdot|s,a) = \hat Z_\theta(s,a)+Q(s,a)$ is trained by VDAL to satisfy \eqref{eq:bellmanDistEqn}.
We believe other distributional methods may benefit from this setting. 

\section{Multivariate Rewards}\label{sec:multivariate_rewards}

The equivalence between DiRL and GANs proposed in the previous section allows us to train deep generative models for approximating the value distribution. One advantage of this approach is when the reward is a \emph{vector}, requiring learning a multivariate distribution for the value function. Multivariate rewards are important when the natural objective in the decision problem cannot be easily composed into a single scalar quantity~\citep{mannor2004geometric,vamplew2011empirical}, and where different objectives may need to be balanced. Examples include RL for clinical trials~\citep{zhao2009reinforcement}, but also learning auxiliary predictions within a standard RL problem~\citep{sutton2011horde,dosovitskiy2016learning}. 

Since GANs have been shown to generate remarkably convincing high dimensional multivariate signals such as images~\citep{goodfellow2014generative,radford2015unsupervised}, we propose that our GAN approach to DiRL would scale well to the multivariate reward case. 

Consider the case where the reward is a vector $r_{t}=r(s_{t},a_{t},s_{t+1})\in\mathbb{{R}}^{m}$.
Given a fixed policy $\pi$, we define the return vector
\begin{equation*}
Z(s,a)=R(s,a)=\sum_{t=0}^\infty \Gamma^{t}r_{t} \in\mathbb{{R}}^{m} , 
\end{equation*} 
where $s_{0}=s,a_{0}=a$.
$\Gamma=\gamma I_{m}$ is the discount matrix, and the multivariate Bellman equation is 
\begin{equation}
Z(s,a)\overset{\bar{d}_{1}}{=}r_{t}+\Gamma Z(s',a'),\label{eq:multidimBellmanEq}
\end{equation}
where $s',a'$ are drawn from $\mathcal{{P}}(\cdot|s,a)$ and policy
$\pi$, respectively.

We note that the discussion of Section \ref{sec:methods} applies to the multivariate reward case. The only modification is that in this case, the generator output and discriminator input are $m$-dimensional vectors.

\section{Wasserstein-distance motivated exploration}

In this section, we propose to use our GAN-based DiRL method for exploration in MDPs. 
In exploration, the agent attempts to reach state-space regions where 
knowledge about the environment or reward function is insufficient,
thereby gaining better experience for future action learning. 

The inability to predict the expected rewards or transitions is generally
a result of one of three factors. (1) Inherent uncertainty of the environment, (2) Inadequate learning model, or (3) Poor experience. The first factor is directly related to the structure of the environment, and we assume that it is beyond our control. The second factor can be mitigated by using a richer class of models, e.g., expressive models such as deep NNs. Here, we focus on the third factor, and propose a method for directing the agent to areas in state space where its past experience is not sufficient to accurately predict the reward and transitions of the state. 

The key feature of our approach to exploration is the following. Assume we have a model of $Z(s,a)$ (namely a model for the distribution of $Z(s,a)$), trained
to minimize $D(Z(s,a),T^{\pi}Z(s,a))$ for some distributional distance
$D$. In regions where we have rich enough experience, the approximated distance 
$D(Z(s,a),T^{\pi}Z(s,a))$ will be small, even though $Z(s,a)$ itself
may have very high moments. Regions where the drawn rewards are statistically
different from past experience will be characterized by a higher discrepancy.
As a result, distributional learning may help us guide exploration
to state-action regions where more samples are required.

We now propose an exploration method based on DiRL. 
Recall that every generator
training step involves an update of the NN parameters $\theta\rightarrow\theta'$ according
to a sampled batch of data. We measure the effect on the distributional model by
inspecting the Wasserstein-$1$ distance, $W_{1}(G_{\theta}(\cdot|s_{t},a_{t}),G_{\theta'}(\cdot|s_{t},a_{t}))$.\footnote{ Although many distributional metrics may be considered, corresponding to the methods suggested in Section \ref{sec:methods}, here we use the Wasserstein-$1$ distance.} 
Following \citet{arjovsky2017wasserstein}, we assume $G_{\theta}(z|s,a)$ to be locally Lipschitz in $(z,\theta)$ with constants $L(z,\theta|s,a)$, such that for all $(s,a)$,
\begin{equation}
\mathbb{E}_{z \sim P_z}L(z,\theta|s,a)=L(\theta|s,a)<\infty.
\end{equation}
\cite{arjovsky2017wasserstein} showed that for any given $(\theta,s,a)$, there exists a neighborhood of $\theta$, denoted $U_{s,a}(\theta)$,  such that for all $\theta'\in U_{s,a}(\theta)$ we have
\begin{equation} \label{eq:W1:exploration::update_bound}
W_{1}(G_{\theta}(\cdot|s,a),G_{\theta'}(\cdot|s,a))\leq L(\theta|s,a)\|\theta-\theta'\|.
\end{equation}

That is, for small enough updates, Equation \eqref{eq:W1:exploration::update_bound} bounds the distributional change of the generator output for each $(s,a)$ by terms of the parameter difference.
Using a gradient-descent method implies an update of the parameters that is proportional to the gradient,
\begin{equation}
\theta-\theta'\propto\nabla_{\theta}\mathcal{{L}}_{\pi}(G_{\theta},f),
\end{equation}
where 
\begin{equation}
\nabla_{\theta}\mathcal{{L}}_{\pi}(G_{\theta},f)=\nabla_{\theta}\mathbb{{E}}_{z\sim P_{z},s'\sim\mathcal{{P}}(\cdot|s,a),a'\sim\pi(\cdot|s')}\Lambda(G_{\theta},D_{\omega})|_{(z,s,a,r,s',a')}.
\end{equation}
Practically, we approximate over a batch of samples. Based on \eqref{eq:Kantor}-\eqref{eq:Lambda-def}, we have that
\begin{equation}
\nabla_{\theta}\mathcal{{L}}_{\pi}(G_{\theta},f)\approx\hat{\mathbb{{E}}}_{z\sim P_{z},s'\sim\mathcal{{P}}(\cdot|s,a),a'\sim\pi(\cdot|s')}\nabla_{\theta}\Lambda(G_{\theta},D_{\omega})|_{(z,s,a,r,s',a')},
\end{equation}
where by $\hat{\mathbb{{E}}}$ we denote the empirical mean over a batch.
Our idea is that the gradient $\nabla_{\theta}\Lambda(G_{\theta},D_{\omega})|_{(z,s,a,r,s',a')}$ is, in effect, a measure of the error in predicting the return distribution at the state-action tuple $s,a$. Thus, we propose to use the magnitude of the gradient as an intrinsic reward function (cf. Section \ref{ssec:exploration_background}). We introduce the combined reward function:
\begin{subequations} \label{eq:exploration}
\begin{align}
\hat{r}(s_{t},a_{t},s_{t+1}) & =r(s_{t},a_{t},s_{t+1})+\eta r^{i}(s_{t},a_{t},s_{t+1}),\quad \eta\geq0,\label{eq:exploration::combinedReward}\\
r^{i}(s_{t,}a_{t},s_{t+1}) & =\|\mathbb{{E}}_{z\sim P_{z},a_{t+1}\sim\pi(\cdot|s_{t+1})}\nabla_{\theta}\varLambda(G_{\theta},D_{\omega})|_{(z,s_{t},a_{t},r_{t},s_{t+1},a_{t+1})}\|.\label{eq:w1me::internalReward}
\end{align}\end{subequations}
Based on this definition, we introduce the Distributional Discrepancy
Motivated Exploration (W-1ME) method, described in Algorithm \ref{alg:Distributional-Distance-Minimizi}.

\subsection{Reward-systematic exploration}\label{sec:Multi-dimensional-distribution-a}

In domains where the reward is sparse, uncertainty in the value distribution may not be an informative signal for exploration.
To overcome sparsity, several exploration methods enrich the reward
signal using some `intrinsic' reward, based on informative measures
of transitions {(}see e.g. \citet{houthooft2016vime} {)}. Following the ideas of model-based exploration~\citep{houthooft2016vime}, we propose to learn the transition probability $\mathcal{{P}}(\cdot|s,a)$, together
with the reward distribution. 

Our main insight is that, by adopting the multivariate reward framework of Section \ref{sec:multivariate_rewards}, we can unify learning the distribution of return and transitions under a single GAN framework. 
Assume that $\mathcal{{S}\subseteq\mathbb{{R}}}^{n}$
has a $L_{2}$ (or equivalent) norm. Then, learning a dynamic model is a special
case of \ref{eq:multidimBellmanEq}, where
\begin{equation}
\tilde{r}(s,a,s')=\left(\begin{array}{c}
r(s,a)\\
s'
\end{array}\right)\label{eq:mulrivar::systematic_rwvector_def}
\end{equation}
is the reward vector, and 
\begin{equation}
\Gamma=\left(\begin{array}{cc}
\gamma I_{m} & 0\\
0 & 0_{n\times n}
\end{array}\right).
\end{equation}

Thus, by simply adding the state to the NNs in the GAN generator and discriminator, and setting the discount factor appropriately, we obtain a GAN that predicts the joint distribution of the return together with the state transition. Note that, since we are learning \emph{joint distributions}, any dependency between the return and state transition should be learned by such a GAN. In addition, the intrinsic reward in Equation \eqref{eq:exploration::combinedReward} can be immediately applied to this GAN, and will in this case include a reward bonus for errors both in the return distribution and in the state transition. 

\begin{algorithm}[h]
\caption{\label{alg:Distributional-Distance-Minimizi}Distributional Discrepancy
Motivated Exploration (W-1ME) }

INPUT: initial policy $\pi(\cdot|s)$, trained model $(G_{\theta},D_{\omega})$. 

PARAMETERS: number of steps $T$, number of noise samples $N_{explore}$, trade-off parameter $\eta$.
\begin{enumerate}
\item For T steps, act according to $a_{t}\sim\pi(\cdot|s_{t})$.
\begin{enumerate}
\item Observe $(s_{t},a_{t},s_{t+1})$ and draw reward $r_{t}$.
\item Draw $N_{explore}$ noise samples $z^{(i)}\sim\mathcal{{N}}(0,I)$
and actions $a_{t+1}^{(i)}\sim\pi(\cdot|s_{t+1})$ .
\item Approximate \eqref{eq:w1me::internalReward} by the empirical mean over
sample $(s_{t},a_{t},r_{t},s_{t+1})$: 
\[
r^{i}(s_{t,}a_{t},s_{t+1})=\left\|\frac{1}{N_{explore}}\sum_{i=1}^{N_{explore}}\nabla_{\theta}\Lambda(G_{\theta},D_{\omega})_{(z^{(i)},s_{t},a_{t},r_{t},s_{t+1},a_{t+1}^{(i)})}\right\|.
\]

\item construct combined rewards $\hat{r}(s_{t},a_{t},s_{t+1}),$ applying
\ref{eq:exploration::combinedReward} using parameter $\eta$.
\end{enumerate}
\item Update policy $\pi$ using rewards $\hat{r}(s_{t},a_{t},s_{t+1})$
and any standard RL method.
\item Train $(G_{\theta},D_{\omega})$ using stored $(s_{t},a_{t},r_{t},s_{t+1})$,
$\pi$ and Algorithm \ref{alg:Distributional-Value-Learn}.
\end{enumerate}
\end{algorithm}

\section{experimental results}

In this section we demonstrate our methods empirically. Our experiments are designed to address the following questions.
\begin{enumerate}
\item Can the VDAL algorithm learn accurate distributions of multivariate returns?
\item Does the W-1ME algorithm result in effective exploration?
\end{enumerate}

\subsection{Multivariate policy evaluation}

Here, we want to demonstrate that our GAN-based method can learn return distributions of multivariate returns, with non-trivial correlation structure between the return variables. For this task, we designed a toy domain where the correlation structure is easy to visualize. 

\begin{figure}
\begin{tabular}{cc}
\includegraphics[trim=0 0 0 14mm,clip,scale=0.4]{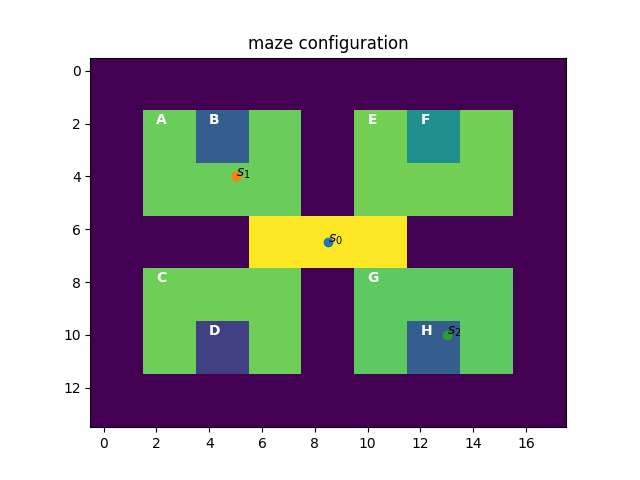} & \includegraphics[scale=0.4]{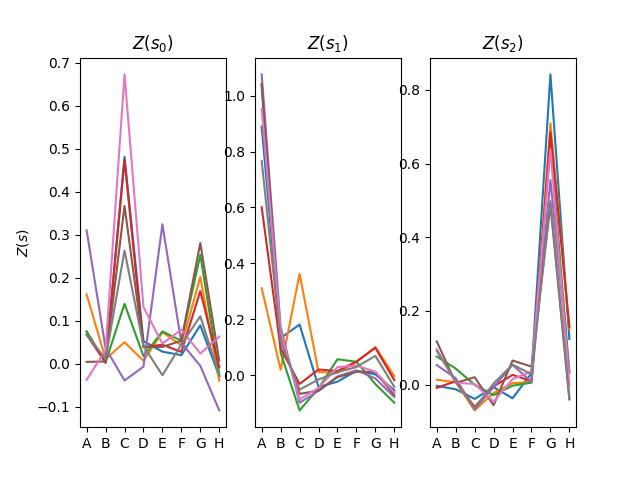}\tabularnewline
(a) & (b)\tabularnewline
\end{tabular}

\caption{ \label{fig::4maze_8rw:Random-policy-evaluation} Random policy evaluation
on multi-reward maze. (a) maze configuration; (b) generated $Z(s)$
samples at different locations, where each line represents a sample in $\mathbb{R}^8$. Note that when starting at the center of the maze ($s_0$), the distribution contains several modes, where in each mode only two reward types have a high return. Conversely, when starting at $s_1$, only returns of type $A$ and $B$ are observed.}
\end{figure}

We consider a multi-reward maze with
a tabular state space -- the Four-room maze in Fig. \ref{fig::4maze_8rw:Random-policy-evaluation}(a). The four allowed actions are moving one unit $UP$, $DOWN$, $LEFT$ or $RIGHT$.
We have 8 reward types corresponding to the subsets $A,\dots,H\subseteq\mathcal{{S}}$, where in each subset of the state space only one type of reward is active, 
\begin{equation}
r_{t}(s_{t},a_{t},s_{t+1})=\frac{1}{1-\gamma}(\delta_{A}(s_{t+1}),\dots,\delta_{H}(s_{t+1}))^{T}\in\mathbb{{R}}^{8}.
\end{equation}
Each room contains a pair of reward types, while the aisle is neutral. 
Thus, when exploring each room, only two reward types can be collected. We consider an agent that moves randomly in the maze. Because of the maze structure, which contains narrow passages between the rooms, the agent is likely to spend most of its time in a single room, and we therefore expect the return distribution to contain different \emph{modes}, where in each mode only two rewards are dominant.

We trained the VDAL algorithm for $1500$ episodes of $350$ steps,
and sampled $Z(s,a)$ at three different states ($s_{0},s_{1},s_{2}$). Action $a$ was randomly chosen at each sample, where we denote the obtained value distribution $Z(s)\triangleq Z(s,a), \ a\sim U(\mathcal{A})$. 

Figure \ref{fig::4maze_8rw:Random-policy-evaluation}(b) shows that
at each location, $Z(s)$ generated higher probability of values for near-by
reward types; while samples from $Z(s_{1})$ tends to present higher
$A$ and $B$ rewards, $Z(s_{2})$ predicts higher $G$'s and $H$'s. When starting at the center of the maze, the return distribution $Z(s_{0})$ is indeed multi-modal, with each mode showing high returns for two different reward types, corresponding to one of the four rooms.

We emphasize that previous DiRL algorithms, such as C-$51$~\citep{bellemare2017distributional}, or quantile regression DQN~\citep{dabney2017distributional}, which rely on a discretization of the value distribution, cannot be used to practically estimate an $8$-dimensional return vector. As our results show, our GAN approach can effectively handle multivariate, correlated, return distributions.

\subsection{DiRL-Based Exploration}

To evaluate the W-1ME exploration algorithm, we propose the 2-Face Climber testbench. In 2-Face Climber (Fig \ref{fig:climber}; see full description in Appendix \ref{sec:climber}), a climber is about to conquer the
summit of a mountain, and there are two possible ways to reach the top.
The South Face is mild and easy to climb. The North face is slippery and much harder to climb, but the route is shorter, and reaching the top bears a greater reward. The climb starts at camp ($s_{0}=0$), where the climber chooses the face to climb. Then, at each step, there are two possible actions. One action progresses the climber towards the summit, while the effect of the other action depends on the mountain face. On the South Side, it causes the climber to slip, and move back to the previous state, while on the North Side, with some probability it can also cause her to fall several steps, or even restart at base-camp. The idea here is that once the model is known, it is easy to always take actions that progress towards the summit, and in this case the north side should be preferred. During exploration, however, the south side is more forgiving when taking a wrong action.

\begin{figure}\label{fig:climber}
\includegraphics[width=0.4\textwidth]{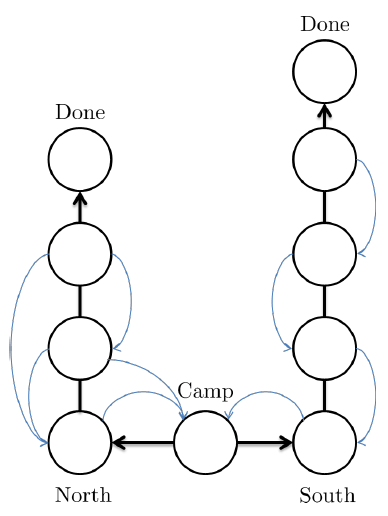}\includegraphics[width=0.6\textwidth]{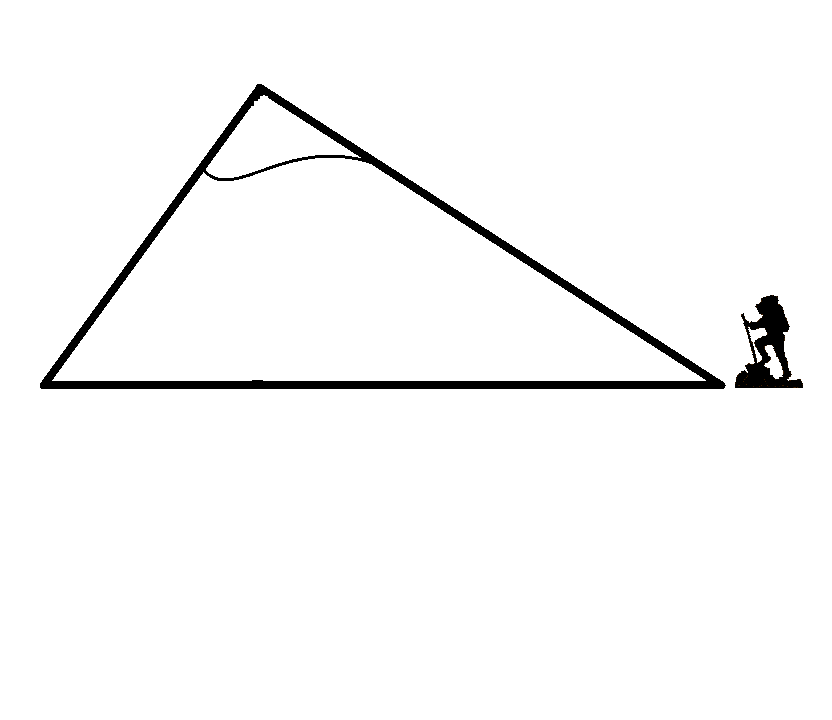}\caption{2-Face Climber. The South Face is easy to climb. The North face is harder to climb, but the route is shorter. Choosing the right action progresses the climber towards the summit (bold edges), while the other action causes her to slip with some probability (light edges).}
\end{figure}

\begin{figure}
\begin{centering}
\emph{\includegraphics[trim=0 0 0 8mm,clip,width=0.5\textwidth]{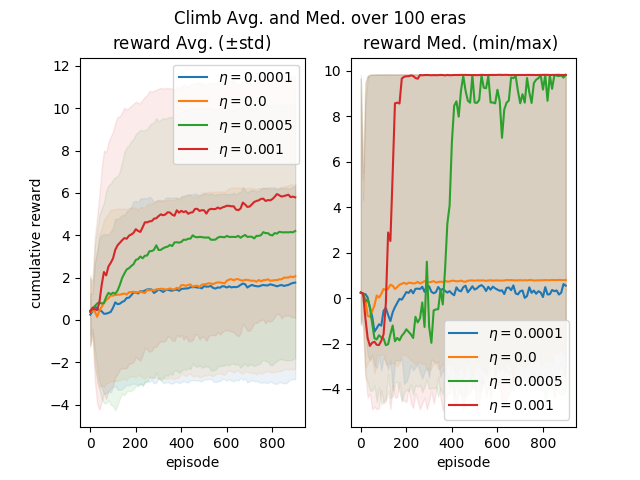}}~
\includegraphics[trim=0 0 0 0mm,clip,width=0.49\textwidth]{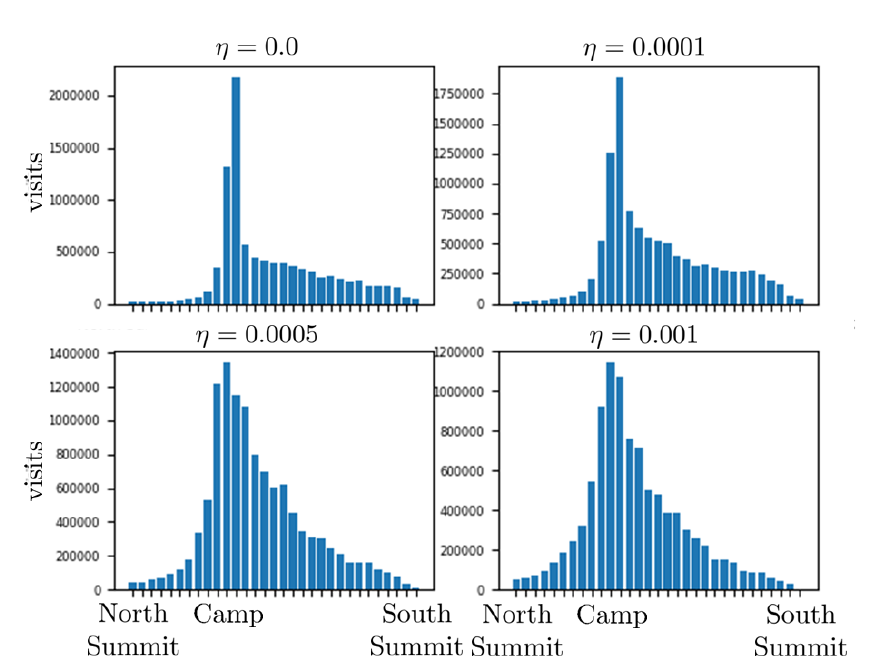}
\par\end{centering}
\centering{}\caption{\label{fig:DQN+W1ME-on-2Face}W-1ME exploration on 2Face climber. \textbf{Left:} Improvement in average and median return using W-1ME+DQN with different $\eta$ parameters. \textbf{Right:} Histograms present the number of visits to each state. Observe that higher $\eta$'s incremented the visit rate to the North face states, resulting in higher average returns. This shows the utility of our exploration method.} 
\label{fig:2Fclimb:State-exploration}

\end{figure}

Figure \ref{fig:DQN+W1ME-on-2Face} shows results of W-1ME exploration using different values of $\eta$ in \eqref{eq:exploration::combinedReward}, where we used reward-systematic exploration (Sec. \ref{sec:Multi-dimensional-distribution-a}). For policy improvement in Algorithm \ref{alg:Distributional-Distance-Minimizi} we used DQN, where we refer to the full algorithm as W-1ME+DQN.
$\eta=0$ is for DQN, where we applied an $\epsilon$-greedy exploration
($\epsilon=0.05$). Average and median are over $100$ independent seeds. Figure \ref{fig:2Fclimb:State-exploration} also presents the
state-space visit counts for the different $\eta$'s. Here we can see that indeed, higher exploration $\eta$'s increment the visit rate to the North face states, resulting in higher average returns.

\section{Conclusion and Outlook}

In this work we showed an interesting equivalence between the distributional Bellman equation and GANs. Based on this equivalence, we proposed a GAN based algorithm for DiRL, which can handle DiRL with high-dimensional, multivariate rewards. We also showed that the multivariate reward formulation allows to unify learning of return and next state distributions, and we proposed a novel exploration method based on this idea, where the prediction error in both return and next state distribution is used as an intrinsic reward. We empirically validated our methods in several RL domains.

Our work paves the way for future investigation of a distributional approach to multi-objective RL~\citep{vamplew2011empirical}. Such would require a distributional \emph{policy optimization} algorithm that can exploit the multi-variate reward distribution from the Bellman GAN. Our unified method for learning state and value distributions also suggests a new direction for model-based RL in high-dimensional state spaces.

\bibliography{references}
\bibliographystyle{iclr2018_conference}

\newpage 
\appendix
\section{The Wasserstein-1 distance}
Let $\mathcal{X}$ be a Polish space with a complete metric $d$, and let
$\mathfrak{B}(\mathcal{X})\subset2^{\mathcal{{X}}}$ denote the $\sigma$-algebra
of Borel subsets. Denote by $\mathcal{{P}}(\mathcal{X})$ the set
of probability measures on $(\mathcal{X},\mathfrak{B}(\mathcal{X}))$.
For $\mu_{1},\mu_{2}\in\mathcal{{P}}(\mathcal{X})$, $\Pi(\mu_{1},\mu_{2})$
is the set of joint distributions whose marginal distributions correspond
to $\mu_{1},\mu_{2}$.
Let $p\in[1,\infty)$. The Wasserstein-p distance w.r.t. the metric
$d$ is defined by

\begin{equation}
W_{p}(\mu_{1},\mu_{2})=(\inf_{\gamma\in\Pi(\mu_{1},\mu_{2})}\mathbb{{E}}_{(X,Y)\sim\gamma}d^{p}(X,Y))^{\frac{1}{p}}.
\end{equation}
An important special case is the \textit{Earth Mover's distance,}
also commonly called the Kantorovich\textendash Rubinstein distance
or simply Wasserstein-$1$ \citep{villani2008optimal}, where $\mathcal{X}=\mathbb{{R}}^{n}$
and

\begin{equation}
W_{1}(\mu_{1},\mu_{2})=\inf_{\gamma\in\Pi(\mu_{1},\mu_{2})}\mathbb{{E}}_{(X,Y)\sim\gamma}\|X-Y\|.
\end{equation}

The Wasserstein-$1$ distance has the following duality property \citep{villani2008optimal}. For any $\mu_{1},\mu_{2}\in\mathcal{{P}}(\mathcal{X})$
with $\int_{\mathcal{{X}}}d(x_{0},x)d\mu_{i}<\infty,\,i=1,2$ (here
$x_{0}$ is an arbitrary point), the $W_{1}$ distance has the following
dual integral probability metric (IPM; \citet{muller1997integral})
form 

\begin{equation}
W_{1}(\mu_{1},\mu_{2})=\sup_{f\in 1-\mathrm{Lip}}\left\{ \int_{\mathcal{{X}}}f(x)d\mu_{1}-\int_{\mathcal{{X}}}f(x)d\mu_{2}\right\} ,\label{eq:App::Wasserstein::duality}
\end{equation}
where $1-\mathrm{Lip}$ is the class of Lipschitz functions $f: \mathcal{X} \rightarrow \mathbb{R}$, with a best admissible Lipschitz constant smaller or equal to $1$.
\section{2-Face Climber testbench}\label{sec:climber}

Consider the following problem: A climber is about to conquer the
summit of a mountain. There are two possible ways to reach the top.
South Face is mild and easy to climb. The north face is steep and much
harder to climb, but the track is shorter, and reaching the top bears a
greater reward.

The climb starts at base-camp ($s_{0}=0$), where the climber chooses the
face to climb by taking an action $a_{0}\in\left\{ {North,South}\right\} $.
When simulation starts, 2 random bit strings are chosen (a string
for each face, 1 digit for every possible state). By $seq(s|face)$
we denote the bit chosen for state $s$ on each face .'Climbing' a
face is made by taking an action $a_{t}\in\left\{ 0,1\right\} $ and
comparing to the digit of current state. We can write transition rule
for $s_{t}\neq0$ as

\begin{equation}
\begin{cases}
\left(\begin{array}{c}
s_{t+1}\\
face_{t+1}
\end{array}\right)=\left(\begin{array}{c}
s_{t}+1\\
face_{t}
\end{array}\right), & a_{t}=seq(s_{t}|face_{t})\\
\left(\begin{array}{c}
s_{t+1}\\
face_{t+1}
\end{array}\right)=\left(\begin{array}{c}
(s_{t}-fall)\vee0\\
face_{t}
\end{array}\right), & a_{t}\neq seq(s_{t}|face_{t}),\,fall\sim unif\{0,\dots,slope(face_{t})\}
\end{cases}
\end{equation}

where $face_{t}\in\{North,South\}$. For $s_{t}=0$ (Camp) we have

\begin{equation}
\begin{cases}
\left(\begin{array}{c}
s_{t+1}\\
face_{t+1}
\end{array}\right)=\left(\begin{array}{c}
1\\
South
\end{array}\right), & a_{t}=1\\
\left(\begin{array}{c}
s_{t+1}\\
face_{t+1}
\end{array}\right)=\left(\begin{array}{c}
1\\
North
\end{array}\right), & a_{t}=0
\end{cases}.
\end{equation}

Simulation is terminated when reaching the top, i.e.

\begin{equation}
s_{t+1}=s_{terminal}(face_{t}).
\end{equation}

\subsection{Rewards}

We have a negative reward (cost) for every climb step (regardless
of action),

\[
r(s_{t},a_{t},s_{t+1},face_{t+1})=C_{face}<0,\,s_{t+1}\neq s_{terminal}(face_{t}),
\]

where 'climbing' the North face typically costs a little more than
the South.

We also have a reward for reaching the top,

\[
r(s_{t},a_{t},s_{t+1},face_{t+1})=R_{face}\gg0,\,s_{t+1}=s_{terminal}(face_{t}).
\]

\subsection{Parametric setup}

We set $s_{terminal}(North)=10,\ s_{terminal}(South)=20,\ slope(North)=4,\ slope(South)=1,\ C_{North}=-0.02,\ C_{South}=-0.01,\ R_{North}=10,\ R_{South}=1$.

\section{Implementation}

\begin{figure}
\begin{tabular}{cc}
\includegraphics[scale=0.8]{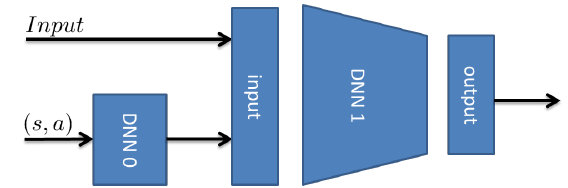} & \includegraphics[scale=0.9]{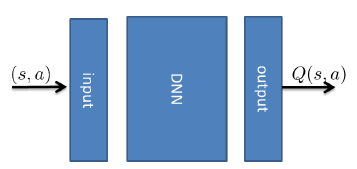}\tabularnewline
(a) & (b)\tabularnewline
\end{tabular}\caption{NN architecture \label{fig:Discriminator-and-Generator_Arch}\label{fig:DQN-net-architecture}}
(a) discriminator and generator, and (b) DQN
\end{figure}

We use the following architecture for both generator and discriminator
(Figure \ref{fig:Discriminator-and-Generator_Arch}(a)). $DNN_{0}$
and $DNN_{1}$ are constructed by a sequence of fully connected linear
layers followed by Leaky ReLU activation. The generator's input is
a Normal-distributed noise, and output dimension is the same as the
return vector. Discriminator output is 1-dimensional. We train \eqref{eq:bellGAN::gameFormula}
using the Adam optimizer.

We implemented DQN using the Double DQN algorithm \citep{van2016deep},
based on an action and a target network. Networks are implemented
using the same architecture (Figure \ref{fig:DQN-net-architecture}(b)).
$DNN$ is constructed by a sequence of fully connected linear layers
followed by ReLU activation. We train DQN using the Adam optimizer with
the Huber loss \citep{mnih2015human}.

In Adam optimizers~\citep{kingma2014adam} we used hyperparameters $\beta_1 = 0.9,\beta_2=0.999$.

In the policy evaluation scenario (Multi-reward maze) we used DQN as part of generator structure \eqref{sec::ZQsimEstim}, where in the exploration (Climber) scenario it was used for policy improvements in Algorithm \ref{alg:Distributional-Distance-Minimizi}. 

Test-specific parameters are stated below. 

\subsection{Multi-reward maze}
\begin{itemize}
\item DDQN:
\begin{itemize}
\item Layers size: input {[}16{]}, DNN {[}16,16{]}, output {[}1{]}.
\end{itemize}
\item Generator:
\begin{itemize}
\item Layers size: DNN0 {[}8,8,8{]} , input {[}128{]}, DNN1 {[}128{]}, output
{[}8{]}.
\item Activation function: Leaky ReLU, Output activation: Linear.
\item Input noise dim: 8.
\end{itemize}
\item Discriminator:
\begin{itemize}
\item Layers size: DNN0 {[}8,8,8{]} , input {[}256{]}, DNN1 {[}128{]}, output
{[}1{]}.
\item Activation function: Leaky ReLU, Output activation: Linear.
\end{itemize}
\item Train parameters: $\lambda=0.1,$ $\gamma=0.95$, minibatch size $64$, learning rate $0.001$.
\end{itemize}

\subsection{2-Face Climber}
\begin{itemize}
\item DDQN:
\begin{itemize}
\item Layers size: input {[}16{]}, DNN {[}16,16{]}, output {[}1{]}.
\end{itemize}
\item Generator:
\begin{itemize}
\item Layers size: DNN0 {[}4,4,4{]} , input {[}128{]}, DNN1 {[}64{]}, output
{[}2{]}.
\item Activation function: Leaky ReLU, Output activation: Linear.
\item Input noise dim: 2.
\end{itemize}
\item Discriminator
\begin{itemize}
\item Layers size: DNN0 {[}4,4,4{]} , input {[}256{]}, DNN1 {[}256,16{]},
output {[}1{]}.
\item Activation function: Leaky ReLU, Output activation: Linear.
\end{itemize}
\item Train parameters: $\lambda=0.1,$ $\gamma=0.9$, minibatch size $64$, learning rate $0.0001$.
\item Exploration parameters: $N_{explore}=4$, $T=32$.
\end{itemize}

\end{document}